\documentclass[conference]{IEEEtran}
\IEEEoverridecommandlockouts

\usepackage{mwe}
\usepackage{multirow}
\usepackage{cite}
\usepackage{amsmath,amssymb,amsfonts}
\usepackage{algorithmic}
\usepackage{graphicx}
\usepackage{textcomp}
\usepackage{xcolor}
\def\BibTeX{{\rm B\kern-.05em{\sc i\kern-.025em b}\kern-.08em
    T\kern-.1667em\lower.7ex\hbox{E}\kern-.125emX}}
\usepackage{url}
    
\begin{document}

\title{Improving Greenland Bed Topography Mapping with Uncertainty-Aware Graph Learning on Sparse Radar Data}

\author{\IEEEauthorblockN{Bayu Adhi Tama$^{1}$, Homayra Alam$^{2}$, Mostafa Cham$^{2}$, Omar Faruque$^{2}$, Jianwu Wang$^{2}$, and Vandana Janeja$^{2}$}
\IEEEauthorblockA{$^{1}$iHARP, University of Maryland Baltimore County (UMBC)\\
$^{2}$Department of Information Systems, University of Maryland Baltimore County (UMBC)\\
\{\textit{bayu; halam3; mcham2; omarf1; jianwu; vjaneja}\}\textit{@umbc.edu}}
}

\maketitle

\begin{abstract}
Accurate mapping of Greenland’s subglacial bed topography is critical for predicting future sea-level rise, yet direct radar measurements are sparse, unevenly distributed, and uncertain. Existing interpolation and modeling approaches either rely heavily on assumptions or struggle to generalize in under-observed regions. We present GraphTopoNet, a graph learning framework designed to integrate heterogeneous supervision sources and incorporates uncertainty-aware regularization via Monte Carlo dropout to improve robustness in data-sparse regions. GraphTopoNet constructs spatial graphs from surface observables (e.g., elevation, velocity, mass balance) and augments them with gradient and polynomial trend components to capture local variability and large-scale structure. To handle data sparsity, we introduce a hybrid loss that combines confidence-weighted radar supervision, dynamically balanced through uncertainty-aware regularization. Applied to three sub-regions of Greenland, GraphTopoNet consistently outperforms interpolation, convolutional, and graph-based baselines, reducing error by up to 60\% while preserving fine-scale glacial features. Beyond methodological gains, the framework addresses an operational need: producing reliable bed maps that support government and scientific agencies in climate forecasting and policy. The approach demonstrates how graph-based machine learning can transform sparse and uncertain geophysical observations into actionable knowledge for large-scale environmental decision-making.

\end{abstract}


\section{Introduction}
The Greenland Ice Sheet plays a critical role in global sea-level rise, governed by complex interactions among ice flow, basal topography, and surface mass balance. Accurate reconstruction of subglacial bed topography is essential for modeling ice dynamics and forecasting future change~\cite{ipcc2023}. However, direct measurements of bed elevation are limited to sparse and unevenly distributed radar observations, introducing substantial uncertainty into existing models. Traditional interpolation methods—such as mass conservation and geostatistical approaches~\cite{morlighem2011mass,bamber2001new}—have improved large-scale reconstructions but struggle in regions with limited data, failing to capture fine-scale variability.

Graph neural networks (GNNs) have emerged as powerful tools for modeling complex spatial relationships in structured domains~\cite{scarselli2008graph}. In this study, we introduce \textbf{GraphTopoNet}, a generalizable GNN-based framework for high-resolution spatial prediction under sparse and uncertain supervision. Our method constructs a graph from gridded surface-derived variables—such as elevation, velocity, and mass balance—and augments the feature space with local gradients and global trend surfaces to capture multi-scale spatial patterns. A Graph Convolutional Network (GCN)~\cite{kipf2017semisupervised} is used to propagate information across the graph, allowing the model to learn both localized and long-range dependencies.

To address data sparsity and observation noise, we design a hybrid loss function with three key components: (i) confidence-weighted supervision based on radar reliability, (ii) auxiliary guidance from full-coverage reference maps, and (iii) epistemic uncertainty regularization via Monte Carlo Dropout~\cite{srivastava2014dropout,kendall2017uncertainties}. This loss formulation enables the model to prioritize trustworthy regions during training while estimating uncertainty in under-observed areas.

We validate GraphTopoNet on high-resolution glaciological datasets from three sub-regions of the Greenland Ice Sheet. Compared to traditional interpolation methods and recent deep learning baselines, our approach yields superior accuracy, structural fidelity, and spatial generalization—even when tested on regions spatially disjoint from training data.



Unlike prior graph-based interpolation methods that primarily focus on generic spatial learning \cite{appleby2020kriging, li2023rainfall}, our work is motivated by a pressing real-world application: reconstructing Greenland’s subglacial bed topography, a critical input for sea-level rise prediction models. Radar observations are sparse and heterogeneous in quality, while reference maps such as BedMachine \cite{morlighem2017bedmachine} provide full coverage but are uncertain in slow-flowing regions. GraphTopoNet directly addresses this operational challenge by combining (i) \textbf{radar confidence weighting}, (ii) \textbf{auxiliary supervision from reference maps}, and (iii) \textbf{Bayesian uncertainty quantification} in a unified framework. This design enables robust extrapolation across under-observed areas, supporting actionable insights for glaciological modeling and climate risk assessment. While our method is generally applicable to spatial prediction problems with sparse and noisy supervision, this paper emphasizes its value in a high-impact government and scientific context: producing more reliable bed maps of Greenland to inform projections of future sea-level rise.

To the best of our knowledge, GraphTopoNet is the first framework to jointly incorporate GNN-based learning, confidence-weighted supervision, and Bayesian uncertainty modeling for high-resolution spatial prediction over irregularly sampled domains.

\section{Related Work}
Spatial interpolation plays a foundational role in environmental modeling, particularly in estimating values at unmeasured locations from sparse observations. Classical techniques, including deterministic methods like Inverse Distance Weighting (IDW)\cite{li2008review} and Triangulated Irregular Networks (TIN)\cite{feng2024critical}, assume spatial proximity implies similarity. Geostatistical approaches, such as Kriging and its variants~\cite{wackernagel2003ordinary,bamber2013new}, rely on variogram-based models to capture spatial dependence. Despite their long-standing use in geosciences, these methods often suffer from rigid assumptions—such as stationarity and Gaussianity—and are typically limited to scalar outputs derived solely from spatial coordinates. Critically, they do not leverage auxiliary covariates (e.g., surface velocity, elevation), which may offer valuable signals for guiding bed topography inference, especially in data-sparse glacial regions.

To overcome these limitations, recent advances in machine learning have introduced data-driven interpolation techniques that incorporate both spatial structure and auxiliary variables. For instance, Kriging Convolutional Networks (KCN)\cite{appleby2020kriging} blend ideas from Graph Neural Networks (GNNs)\cite{kipf1609semi} with geostatistics to better capture spatial dependencies while preserving inductive capabilities. Similarly, the Graph for Spatial Interpolation (GSI) model~\cite{li2023rainfall} adapts GNNs to learn dynamic spatial relationships without relying on fixed adjacency structures, improving accuracy in irregular and sparse spatial domains such as rainfall estimation. These developments demonstrate the growing utility of graph-based and neural spatial models in addressing the limitations of traditional interpolation.

Estimating subglacial bed topography remains a fundamental challenge in cryosphere research, as it governs basal hydrology and controls glacier flow dynamics. Traditional mass conservation models like BedMachine~\cite{morlighem2017bedmachine} incorporate ice velocity, surface elevation, and surface mass balance to infer the bed, but rely heavily on dense radar data in fast-flowing regions and interpolation elsewhere. More recent work has turned to machine learning and physics-informed frameworks to address limitations in low-data areas.

Cheng et al.\cite{cheng2024forward} proposed a physics-informed neural network (PINN) for Helheim Glacier, integrating conservation of momentum (via the shallow-shelf approximation) directly into the loss function. This allowed interpolation of sparse thickness measurements while preserving physically plausible flow fields. Yi et al.\cite{yi2023evaluating} benchmarked multiple statistical and ML models—e.g., XGBoost and Gaussian Processes—for bed prediction, demonstrating that hybrid models combining learning and geostatistics achieve superior accuracy. Meanwhile, Leong and Horgan~\cite{leong2020deepbedmap} introduced DeepBedMap, a GAN-based model that fuses multiresolution inputs (e.g., velocity, surface elevation) to enhance subglacial terrain reconstruction over Antarctica.

Most closely related to our work is DeepTopoNet~\cite{tama2025deeptoponet}. That framework integrates gradient- and trend-augmented features into a CNN with dynamic radar–BedMachine loss balancing, but remains limited to grid-based local patterns and lacks explicit uncertainty modeling. GraphTopoNet extends this line by leveraging graph-based message passing to propagate information across sparse neighborhoods, introducing uncertainty-aware regularization via Monte Carlo dropout, and validating across multiple Greenland subregions with spatial extrapolation. This progression yields a more generalizable and operationally relevant framework for uncertainty-aware subglacial mapping.

\section{Problem Formulation}
Accurate reconstruction of subglacial bed topography is hindered by the sparsity and uneven distribution of radar-derived measurements, which are also subject to varying levels of observational noise. To address this challenge, we formulate the problem as a supervised graph-based regression task that leverages high-resolution surface observables. Let $X \in \mathbb{R}^{N \times F}$ denote the input feature matrix, where $N$ is the number of spatial grid points and $F$ is the number of surface-derived covariates, including surface mass balance (\texttt{SMB}), surface elevation (\texttt{elv}), horizontal and vertical velocity components (\texttt{v$_{\texttt{x}}$}, v$_{\texttt{y}}$), and surface thickening rate ($\frac{d\texttt{h}}{d\texttt{t}}$). These features are encoded over a spatial graph, where nodes correspond to pixels on the grid and edges capture local spatial adjacency.

The objective is to learn a mapping function $f_{\theta}: X \rightarrow Y$ that predicts the bed elevation $Y \in \mathbb{R}^{N}$ from the input features. We denote the continuous bed topography surface as $b(x, y)$, with $Y$ representing its discretized counterpart over the spatial grid. Supervision is provided by a combination of radar observations and BedMachine-derived reference values. The learning objective is defined by a hybrid loss function that integrates multiple forms of supervision and uncertainty regularization:
\begin{equation}
    \mathcal{L}_{\text{total}} = \mathcal{L}_{\text{radar}} + \mathcal{L}_{\text{ref}} + \mathcal{L}_{\text{uncertainty}},
\end{equation}
where $\mathcal{L}_{\text{radar}}$ penalizes prediction errors at radar-observed locations, weighted by a confidence map; $\mathcal{L}_{\text{ref}}$ applies full-grid supervision using reference BedMachine data; and $\mathcal{L}_{\text{uncertainty}}$ regularizes the model via epistemic uncertainty estimated through Monte Carlo Dropout. This formulation enables spatially coherent, uncertainty-aware inference of subglacial bed topography in regions with heterogeneous data coverage.

\begin{figure*}[ht!]
    \centering
    \includegraphics[width=1\linewidth]{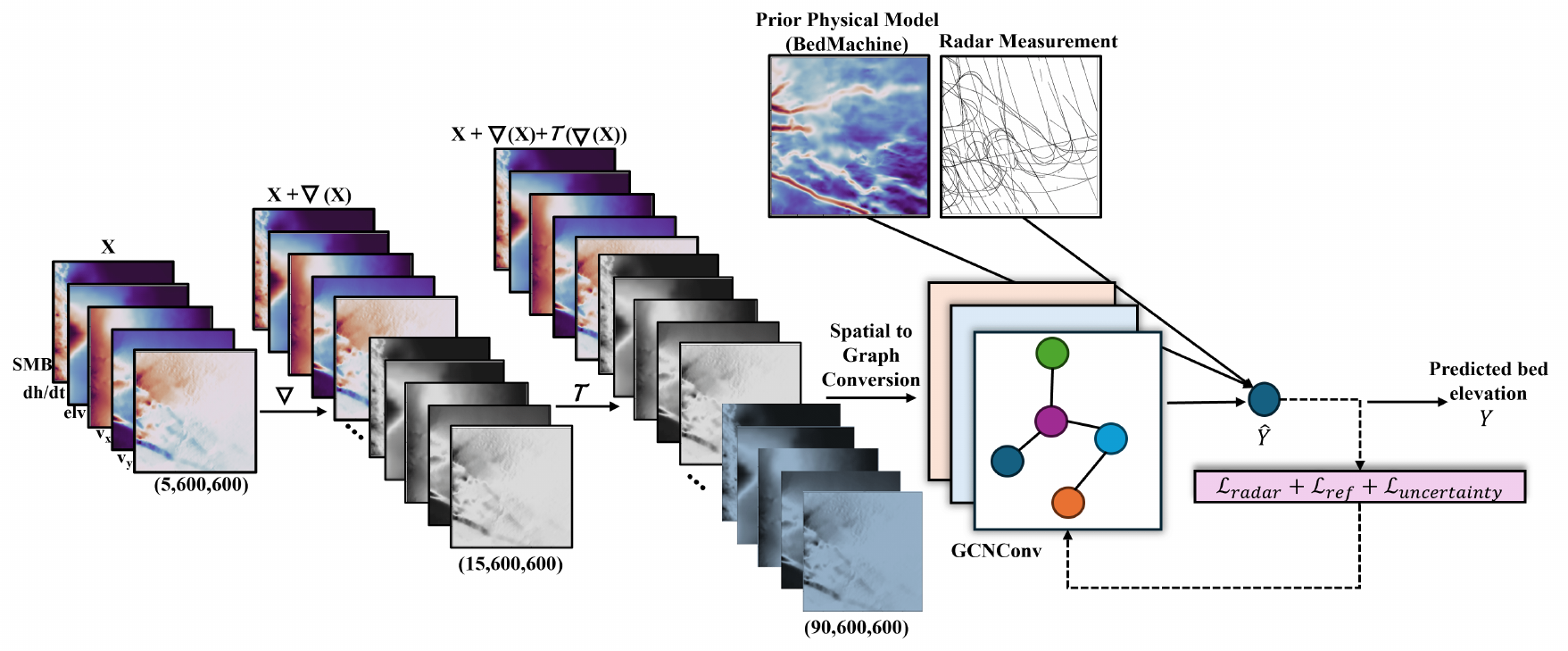}
    \caption{Overview of the proposed GraphTopoNet framework. Surface-derived features (e.g., elevation, velocity, SMB) are processed and encoded over a spatial graph. Gradient and trend surface augmentations enhance the input representation. The graph is passed through a multi-layer Graph Convolutional Network (GCN) to predict bed topography. Supervision is provided by both radar observations and the BedMachine map, with uncertainty-aware loss balancing to guide learning in sparse or noisy regions.}
    \label{framework}
\end{figure*}

\section{Proposed Framework}
The proposed GraphTopoNet framework integrates spatial graph learning, multi-source supervision, and uncertainty quantification to predict subglacial bed topography from surface observations. As illustrated in Figure~\ref{framework}, the pipeline begins with preprocessing and feature augmentation, where gradient and polynomial trend surface features are derived from raw surface variables. These features are mapped to a graph structure, enabling localized message passing through stacked Graph Convolutional layers. The model is trained using a hybrid loss function that incorporates radar-weighted supervision, BedMachine fallback guidance, and epistemic uncertainty regularization. We describe each step in the framework as follows.

\subsection{Prior Physical Model (BedMachine)}
BedMachine~\cite{morlighem2017bedmachine,morlighem2020deep} provides one of the most comprehensive and widely adopted maps of Greenland’s subglacial topography. It utilizes a mass conservation framework in fast-flowing ice regions, integrating radar-derived ice thickness, satellite-derived surface velocity, and surface mass balance estimates to infer bed elevation. The core principle is the conservation of mass, expressed as $\nabla \cdot (s - b) \mathbf{v} = \dot{a}$, where $s$ denotes surface elevation, $b$ is the unknown bed elevation, $\mathbf{v}$ represents ice velocity, and $\dot{a}$ is the apparent mass balance derived from observed thickness change and regional climate models. While BedMachine achieves high fidelity in dynamic regions, its accuracy diminishes in slow-flowing areas where it resorts to spatial interpolation across sparsely sampled radar transects. The resulting product is a seamless bed elevation map at up to 150-meter resolution. In our study, BedMachine serves as a reference dataset for supervised learning. Unlike traditional approaches that depend heavily on mass conservation or interpolation, our proposed \texttt{GraphTopoNet} framework introduces a graph-based deep learning strategy capable of modeling complex spatial dependencies and explicitly quantifying predictive uncertainty across the ice sheet domain.

\subsection{Feature Augmentation}
To enhance model expressiveness and capture spatial heterogeneity, we employ a feature augmentation pipeline that integrates raw spatial covariates with their derived representations. The input features include surface mass balance (\texttt{SMB}), surface elevation (\texttt{elv}), surface thickening rate ($\frac{d\texttt{h}}{d\texttt{t}}$), and velocity components (\texttt{v$_{\texttt{x}}$}, v$_{\texttt{y}}$). Beyond these raw features, we compute spatial gradients ($\nabla$) and polynomial trend surfaces ($\mathcal{T}$) to model both local and global spatial dependencies. The spatial gradients of each feature are calculated to capture local variability. For a spatial field $X \in \mathbb{R}^{H \times W}$, we compute finite-difference approximations of the partial derivatives: $\nabla_x X = \frac{\partial X}{\partial x}$ and $\nabla_y X = \frac{\partial X}{\partial y}$. These gradients reveal the direction and magnitude of local changes. Each feature is thus expanded into a triplet: ${X, \nabla_x X, \nabla_y X}$, resulting in a richer representation of spatial transitions.

To capture large-scale trends, we fit a polynomial surface to each feature. Given spatial coordinates $\mathbf{s} = (x, y)$, we generate a polynomial basis of degree $d$ (e.g., $d=2$), and fit:
\begin{equation}
    \mathcal{T}(X) = \sum_{i+j\leq d} \beta_{ij}x^{i}y^{j}
\end{equation}
where $\beta_{ij}$ are coefficients estimated via least squares. This expansion models global spatial trends and long-range dependencies. The final representation for each input includes the original field, its gradients, and its polynomial trend components. By stacking these representations across all input variables, we obtain the final augmented feature tensor:

\begin{equation}
    X_{\text{final}} = [X,X_{\text{grad}}, X_{\text{trend}}]
\end{equation}
where $X$ denotes the raw features, $X_{\text{grad}}$ the gradient-based features, and $X_{\text{trend}}$ the polynomial trend surfaces. This enriched feature space enables the model to leverage both fine-grained local signals and broad-scale spatial patterns essential for robust topographic inference.

\subsection{Graph Representation of Spatial Grid}
To exploit spatial correlations inherent in the input data, we transform the 2D grid into a graph-based structure compatible with Graph Convolutional Networks (GCNs)~\cite{kipf2017semisupervised}. In this formulation, each pixel $(i, j)$ in the input grid of dimensions $H \times W$ is treated as a graph node, forming the node set $V$. Edges are established between adjacent pixels—both horizontally and vertically—to encode local spatial dependencies.

Specifically, for each node $v_{i,j} \in V$, we construct the undirected edge set $E$ by adding bidirectional edges to its immediate neighbors:
\begin{multline}
E=\{(v_{i,j},v_{i+1,j}),(v_{i+1,j,v_{i,j}}),(v_{i,j},v_{i,j+1}),(v_{i,j+1},v_{i,j})\\\mid 0\leq i < H, 0\leq j <W\}
\end{multline}
This bidirectional edge scheme ensures symmetric information flow between neighboring pixels, which is essential for capturing local context during message passing in GCNs. The graph structure is encoded in a sparse edge index matrix $E_{\text{index}} \in \mathbb{Z}^{2 \times |E|}$, where each column represents a directed edge $(v_{\text{src}}, v_{\text{dst}})$. This efficient representation allows the GCN to operate over irregular spatial domains while preserving the original topological layout of the input grid. Finally, feature vectors associated with each pixel are reshaped into a node feature matrix $X \in \mathbb{R}^{N \times F}$, where $N = H \times W$ is the total number of nodes and $F$ is the number of input features per node. This conversion facilitates the integration of structured spatial priors into the learning process and supports effective graph-based modeling of geospatial data.

\subsection{Graph Learning Architecture}
We design a specialized graph neural network, \texttt{BedTopoGCN}, to model subglacial bed topography using spatially structured inputs. This architecture builds upon Graph Convolutional Networks (GCNs)~\cite{kipf2017semisupervised}, which enable localized message passing across nodes defined on a spatial grid. \texttt{BedTopoGCN} comprises a stack of three GCN layers. The model receives as input a node feature matrix $\mathbf{X} \in \mathbb{R}^{N \times F}$, where each node corresponds to a pixel in the input grid, and an edge index matrix $E_{index}$ that defines adjacency relations among neighboring nodes. The forward pass through the network is given by:
\begin{multline}
H^{(1)}=\text{\texttt{ReLu}}(\text{GCNConv}(X,E_{index})), \\H^{(2)}=\text{\texttt{ReLu}}(\text{GCNConv}(H^{(1)},E_{index})), \\Y=\text{GCNConv}(H^{(2)},E_{index})  
\end{multline}
Here, $Y \in \mathbb{R}^{N \times 1}$ denotes the predicted bed elevation at each node. This layered GCN structure enables the model to aggregate features from increasingly broader spatial neighborhoods, capturing both local and regional topographic patterns. The use of \texttt{ReLu} activations introduces non-linearity between layers, allowing the network to learn complex mappings from surface features to bed elevation. The model is trained end-to-end using a hybrid loss function described in the following section.

\subsection{Hybrid Loss Function}
\label{sec:loss}
To achieve accurate and uncertainty-aware bed topography predictions, we introduce a hybrid loss formulation that combines multiple supervision signals and regularizes the model via epistemic uncertainty. Our loss function consists of three complementary components: (i) radar-supervised loss with confidence weighting, (ii) auxiliary supervision from reference bed topography (e.g., BedMachine), and (iii) epistemic uncertainty regularization derived from stochastic forward passes.

Given the predicted bed elevation $\hat{Y}$, radar-derived ground truth $Y_{\text{radar}}$ with corresponding confidence weights $C_{\text{radar}}$, and the full-grid BedMachine reference $Y_{\text{ref}}$, we define the following loss terms:
\begin{itemize}
\item \texttt{Radar Confidence Loss} ($\mathcal{L}_{\text{radar}}$):
This term enforces accurate predictions at radar-observed locations while incorporating measurement reliability. The squared error is modulated by the radar confidence map:
\begin{equation}
\mathcal{L}_{\text{radar}}=\frac{1}{|R|}\sum_{i\in R}C_{\text{radar},i}\cdot(\hat{Y}_{i}-Y_{\text{radar},i})^{2},    
\end{equation}
where $R$ is the set of radar-labeled indices. The confidence score $C_{\text{radar}, i} \in [0, 1]$ reflects the proximity of pixel $i$ to the nearest radar observation. Specifically, for each grid cell, we compute the Euclidean distance to the closest radar-labeled pixel, assigning a distance of zero for radar-observed points. This distance is then transformed into a soft confidence score using an exponential decay function:
\begin{equation}
    C_{\text{radar},i} = \text{exp}\left(-\frac{d_{i}^{2}}{2\sigma^{2}}\right),
\end{equation}
where $d_i$ is the distance from pixel $i$ to the nearest radar observation and $\sigma$ controls the rate of decay. This formulation encourages the model to prioritize learning in regions close to radar measurements while gracefully attenuating the influence of predictions made further away.

\item \texttt{Reference Bed Loss} ($\mathcal{L}_{\text{ref}}$):  
For regions not covered by radar, we compute the mean squared error with respect to the BedMachine map as a fallback supervision signal:
\[
\mathcal{L}_{\text{ref}} = \frac{1}{|B|} \sum_{i \in B} (\hat{Y}_i - Y_{\text{ref}, i})^2,
\]
where $B$ is the set of non-radar (BedMachine-only) grid cells.
\item \texttt{Epistemic Uncertainty Regularization} ($\mathcal{L}_{\text{uncertainty}}$):  
To encourage the model to express its uncertainty in regions with sparse or unreliable observations, we apply Monte Carlo Dropout~\cite{gal2016dropout} during inference and penalize the predictive variance~\cite{kendall2017uncertainties}:
\[
\mathcal{L}_{\text{uncertainty}} = \frac{1}{N} \sum_{i=1}^{N} \text{Var}(\hat{Y}_i).
\]
\end{itemize}
To dynamically balance these objectives, we adopt an uncertainty-based weighting mechanism. Each loss term $\mathcal{L}_k$ is scaled by a learnable precision parameter $\sigma_k^2$, resulting in the following total loss:
\begin{equation}
\mathcal{L}_{\text{total}}=\sum_{k=1}^{3}\left(\frac{1}{2\sigma_{k}^{2}}\mathcal{L}_{k}+\text{log}\sigma_{k}\right)
\end{equation}
This formulation, implemented via a dedicated \texttt{LossBalancer} module, enables the model to automatically adjust the relative importance of each loss signal during training, improving robustness and convergence across heterogeneous data densities.

This hybrid loss formulation reflects the realities of scientific big data applications. Radar measurements, though highly accurate, are extremely sparse; reference maps such as BedMachine offer spatially complete guidance but are less reliable in certain regions. By combining both sources, GraphTopoNet balances trustworthy local supervision with weak global coverage. The learnable uncertainty-based weighting allows the model to dynamically adjust these contributions during training, ensuring that predictions remain robust in data-rich as well as data-sparse regions. This dual-source strategy mirrors operational practice in government and environmental agencies, where heterogeneous data sources must be reconciled to build actionable decision-support products.

\section{Experiments}
\subsection{Datasets}
\begin{figure}
    \centering
    \includegraphics[width=1\linewidth]{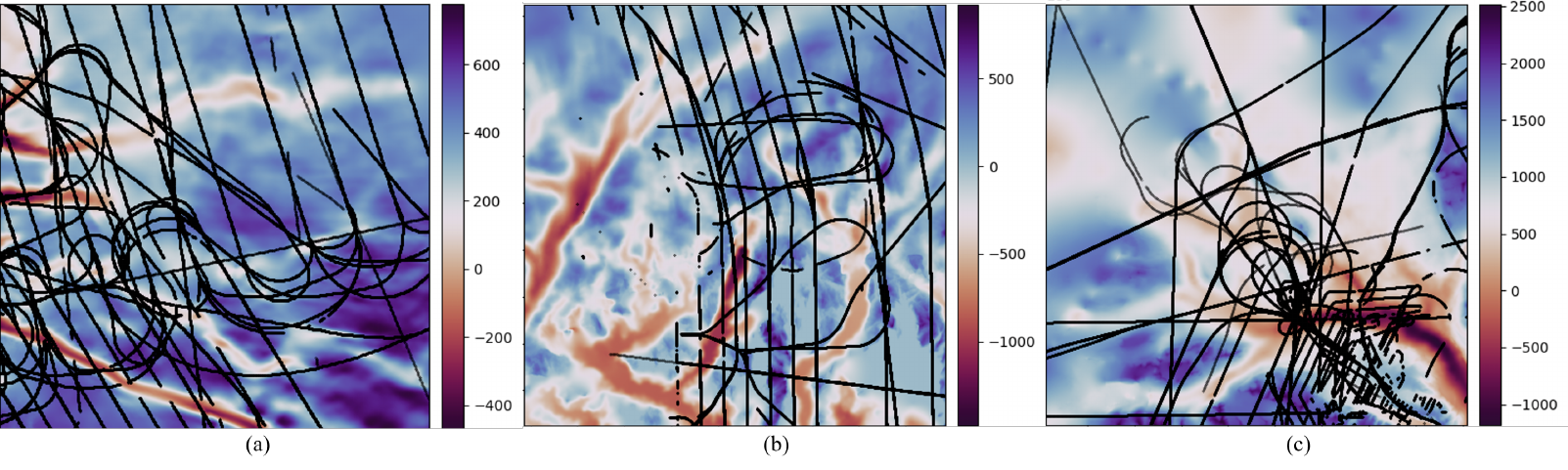}
    \caption{Spatial visualization of BedMachine-derived bed topography with overlaid radar observation points for each sub-region, such as Upernavik Isstr{\o}m (a), Hayes (b), and Kangerlussuaq (c). The figure illustrates the spatial coverage of radar data in relation to the underlying bed elevation.}
    \label{dataset}
\end{figure}

We evaluate our method on three sub-regions of the Greenland Ice Sheet—Upernavik, Hayes, and Kangerlussuaq—each exhibiting distinct glaciological characteristics and varying levels of radar coverage (see Figure~\ref{dataset}). Table~\ref{dataset_summarization} summarizes the bed elevation ranges and the number of radar-derived data points available in each region. These radar picks serve as sparse but high-accuracy ground truth for model supervision. Each sub-region is represented as a $600 \times 600$ gridded domain with a spatial resolution of 150 meters. At each grid point, we extract a set of surface-derived input features that serve as covariates for bed elevation prediction. The features are defined on a uniform spatial grid and sourced from publicly available datasets. 

Each grid cell is associated with a feature vector composed of the following variables (i) surface elevation (\texttt{elv}): Height of the ice sheet surface above a reference level, sourced from ArcticDEM~\cite{howat2014greenland}, surface velocity (\texttt{v$_{\texttt{x}}$}, v$_{\texttt{y}}$): Horizontal components of ice velocity, indicating the direction and speed of ice flow~\cite{joughin2010greenland}, surface mass balance (\texttt{SMB}): Net accumulation or ablation rate, derived from regional climate models~\cite{noel2018modelling,imbie2020mass}, and surface thickening rate ($\frac{d\texttt{h}}{d\texttt{t}}$): Temporal change in ice thickness, reflecting surface elevation change over time~\cite{millan2022ice}. These features provide complementary information about surface processes, which our graph-based model leverages to infer the underlying bed topography. The spatial resolution of the data is consistent across regions, and the features are standardized before model training to ensure numerical stability.

\begin{table}[ht!]
\caption{Summary of the datasets used in this study, including bed height range and radar data point distribution.}
\label{dataset_summarization}
\centering
\resizebox{0.47\textwidth}{!}{%
\begin{tabular}{cccc}
\hline
Sub-regions&Min\_bed\_height&Max\_bed\_height&\#Radar data points\\
\hline\hline
Upernavik&-478.7407&778.9677&316,559\\
Hayes&-1475.9999&922.0001&15,532\\
Kangerlussuaq&-1179.0000&2518.0000&22,055\\
\hline
\end{tabular} 
}
\end{table}

\subsection{Experimental Settings}
To handle large input domains and promote robust training, we adopt a patch-based training strategy on each $600 \times 600$ sub-region. The input surface features are first normalized and augmented with gradient and polynomial trend surface features to enrich spatial context. Each input tensor is then divided into overlapping patches of size $16 \times 16$ using a stride of 8, resulting in dense coverage of the spatial grid. We partition the resulting patch dataset into training and validation subsets using an 80/20 random split. During training, the model is optimized using the Adam optimizer with a learning rate of $1 \times 10^{-4}$ and a batch size of 16. Model performance is evaluated on the validation set at each epoch, and training is terminated early if no improvement is observed in the validation loss for 5000 consecutive epochs (patience = 5000), with a maximum training limit of 20,000 epochs.

To support uncertainty-aware supervision, we perform Monte Carlo Dropout sampling with $N=10$ stochastic forward passes per input during training. The total loss is computed using the hybrid objective described in Section~\ref{sec:loss}, which integrates radar-weighted supervision, fallback BedMachine guidance, and an epistemic uncertainty penalty. The relative importance of each loss component is dynamically balanced using the learnable \texttt{LossBalancer} module. All graph-based computations are implemented using the PyTorch Geometric framework. For each input patch, spatial connectivity is defined via 4-neighbor adjacency, and converted into a graph using the \texttt{grid\_to\_graph} routine. The final model, \texttt{BedTopoGCN}, consists of three GCN layers with 128 hidden units.

\subsection{Evaluation Metrics}
We evaluate model performance using a combination of pointwise error metrics and structural similarity measures to comprehensively assess both accuracy and spatial fidelity of the predicted bed topography. The following metrics are used: (i) Mean Absolute Error (MAE): Measures the average magnitude of absolute prediction errors across all grid cells, (ii) Root Mean Squared Error (RMSE): Emphasizes larger errors by computing the square root of the mean squared differences between predictions and ground truth, (iii) coefficient of determination (R$^2$): Quantifies the proportion of variance in the ground truth explained by the predictions, (iv) Structural Similarity Index Measure (SSIM): Evaluates the perceptual similarity between predicted and reference bed topography maps by comparing local patterns of pixel intensity, and (v) Peak Signal-to-Noise Ratio (PSNR): Assesses the ratio between the maximum possible signal and the noise introduced by prediction errors, reflecting reconstruction quality.

\subsection{Baselines}
To evaluate the effectiveness of our proposed method, we compare its performance against a suite of representative baseline models commonly used in spatial interpolation and geospatial machine learning tasks. These baselines include both traditional non-parametric interpolation techniques and data-driven deep learning models. The traditional methods serve as interpretable, low-complexity benchmarks, while the learning-based models represent state-of-the-art approaches capable of capturing non-linear spatial relationships. We briefly describe the baseline models as follows.
\begin{itemize}
    \item \texttt{Inverse Distance Weighting (IDW)}~\cite{johnston2001using,shepard1968two}: a classical spatial interpolation method that estimates bed elevation at unobserved locations by averaging nearby radar observations, weighted inversely by their Euclidean distance. We use a power parameter of 2 and 4,000 nearest neighbors to ensure smooth, distance-sensitive interpolation across the grid.
    \item \texttt{Nearest Neighbor}~\cite{hart1968condensed}: a simple baseline that assigns bed elevation values to each grid point based on the weighted average of its $k$ nearest radar observations, where inverse distance is used as the weighting function. We set $k = 10{,}000$ to capture sufficient local context across the grid.
    \item \texttt{Gaussian-Smoothed Grid Interpolation (GSGI)}~\cite{mitas1999spatial}: a two-step baseline that first applies linear interpolation via \texttt{griddata} to estimate bed elevation values across the grid and then smooths the result using a Gaussian filter with a fixed kernel width ($\sigma = 5$).
    \item \texttt{MLP}~\cite{schmidhuber2015deep}: a fully connected multi-layer perceptron (MLP) trained on radar-supervised input patches augmented with gradient and polynomial trend features. This non-spatial baseline serves as a strong reference for evaluating the value of spatial inductive biases in topography prediction.
    \item \texttt{MLP-PE}~\cite{li2021learnable}: an enhanced multilayer perceptron that incorporates sinusoidal positional encoding to represent spatial coordinates, allowing the model to better capture location-dependent patterns in bed topography. 
    \item \texttt{MLP-FE}~\cite{tancik2020fourier}: a multilayer perceptron augmented with Fourier feature positional encoding to model high-frequency spatial patterns.
    \item \texttt{U-Net}~\cite{ronneberger2015u}: a convolutional U-Net architecture designed for spatially dense bed elevation prediction, incorporating multi-scale feature extraction and skip connections for accurate reconstruction. 
    \item \texttt{U-Net++}~\cite{zhou2018unet++}: a nested U-Net++ architecture that captures multi-scale spatial dependencies with dense skip connections for refined bed topography prediction. 
    \item \texttt{U-Net3+}~\cite{huang2020unet}: a full-scale skip-connected convolutional architecture that enhances spatial detail recovery by integrating multi-resolution feature maps at each decoder stage. This model is designed to improve fine-grained bed elevation predictions through dense encoder–decoder connectivity.
    \item \texttt{Att. U-Net}~\cite{oktay2018attention}: an encoder–decoder architecture enhanced with attention gates that selectively highlight informative spatial features during decoding. This model improves feature fusion across scales, enabling more precise bed topography predictions in complex terrain.
    \item \texttt{ChebNet}~\cite{defferrard2016convolutional}: a spectral graph convolutional network using Chebyshev polynomial filters to capture localized spatial dependencies over grid-structured data. This model propagates node features through multi-hop neighborhoods without requiring explicit positional encoding.
    \item \texttt{GraphSAGE}~\cite{hamilton2017inductive}: a spatial graph neural network that aggregates neighborhood information through learned functions, enabling localized and scalable learning over irregular spatial domains. This model captures structural dependencies in bed topography by combining node-wise feature encoding with neighborhood context.
    \item \texttt{GAT}~\cite{veličković2018graph}: a graph attention network that dynamically weighs neighboring nodes using attention coefficients, allowing the model to focus on informative spatial relationships. Monte Carlo Dropout is applied to capture epistemic uncertainty, enhancing robustness in regions with sparse or noisy radar data. 
    
\end{itemize}

\begin{table*}[ht!]
\caption{Performance comparison of the proposed model against baseline methods across three sub-regions for bed topography prediction. The best-performing results are highlighted in \textbf{bold}, while the second-best results are underlined.}
\label{result_quantitative}
\centering
\resizebox{1\textwidth}{!}{
\begin{tabular}{l|lllll|lllll|lllll}
\hline
\multirow{2}{*}{Method}&\multicolumn{5}{|c|}{Upernavik}&\multicolumn{5}{|c|}{Hayes}&\multicolumn{5}{|c}{Kangerlussuaq}\\
\cline{2-16}
&MAE$\downarrow$&RMSE$\downarrow$&R$^{2}\uparrow$&SSIM$\uparrow$&PSNR$\uparrow$&MAE$\downarrow$&RMSE$\downarrow$&R$^{2}\uparrow$&SSIM$\uparrow$&PSNR$\uparrow$&MAE$\downarrow$&RMSE$\downarrow$&R$^{2}\uparrow$&SSIM$\uparrow$&PSNR$\uparrow$\\
\hline\hline
IDW~\cite{johnston2001using,shepard1968two}&63.152&101.049&0.653&0.784&21.901&209.486&293.361&0.153&0.277&18.249&196.713&321.180&0.562&0.838&21.222\\
Nearest Neighbor~\cite{hart1968condensed}&76.159&118.260&0.524&0.791&20.535&202.742&280.908&0.223&0.270&18.626&277.796&416.161&0.265&0.803&18.972\\
GSGI~\cite{mitas1999spatial}&58.659&95.680&0.689&0.829&22.375&219.123&314.697&0.026&0.274&17.639&143.804&251.611&0.731&0.895&23.342\\
MLP~\cite{schmidhuber2015deep}&42.631&59.860&0.878&0.804&26.449&102.927&140.338&0.806&0.593&24.654&118.333&156.625&0.896&0.903&27.460\\
MLP-PE~\cite{li2021learnable}&49.707&70.459&0.831&0.695&25.033&142.093&209.626&0.568&0.363&21.168&120.952&159.470&0.892&0.869&27.303\\
MLP-FE~\cite{tancik2020fourier}&45.320&61.598&0.871&0.816&26.200&111.570&150.825&0.776&0.557&24.028&146.770&193.961&0.840&0.893&25.603\\
U-Net~\cite{ronneberger2015u}&268.316&301.553&-2.094&0.254&12.404&288.862&341.714&-0.149&-0.012&16.924&478.111&611.446&-0.587&0.259&15.630\\
U-Net++~\cite{zhou2018unet++}&142.360&169.485&0.023&0.603&17.409&345.313&400.354&-0.577&0.018&15.548&407.466&506.342&-0.088&0.749&17.268\\
U-Net3+~\cite{huang2020unet}&101.697&126.216&0.458&0.713&19.969&306.218&403.146&-0.599&0.165&15.488&249.100&304.241&0.607&0.806&21.693\\
Att. U-Net~\cite{oktay2018attention}&39.230&45.495&0.930&0.918&28.832&73.271&110.245&0.880&0.802&26.750&41.821&54.575&0.987&0.963&36.617\\
ChebNet~\cite{defferrard2016convolutional}&14.567&20.256&0.986&0.945&35.861&31.531&48.827&0.977&0.890&33.824&24.483&35.678&0.995&0.981&40.309\\
GraphSAGE~\cite{hamilton2017inductive}&12.460&17.097&0.990&0.949&37.333&33.281&48.923&0.976&0.878&33.807&\underline{21.805}&\underline{32.619}&\underline{0.996}&\underline{0.982}&\underline{41.088}\\
GAT~\cite{veličković2018graph}&\underline{10.432}&\underline{14.655}&\underline{0.993}&\underline{0.973}&\underline{38.672}&\underline{26.396}&\underline{36.191}&\underline{0.987}&\underline{0.935}&\underline{36.425}&24.166&37.256&0.994&0.981&39.933\\
\textbf{GraphTopoNet}&\textbf{7.328}&\textbf{10.392}&\textbf{0.996}&\textbf{0.979}&\textbf{41.658}&\textbf{19.740}&\textbf{31.530}&\textbf{0.990}&\textbf{0.938}&\textbf{37.623}&\textbf{11.523}&\textbf{18.549}&\textbf{0.999}&\textbf{0.992}&\textbf{45.991}\\
\hline
\end{tabular}
}
\end{table*}

\begin{table*}[ht!]
\caption{Results of the ablation study, assessing the effect of excluding $\nabla$, $\mathcal{T}$, $\nabla$ \& $\mathcal{T}$, and $\mathcal{L}_{\text{ref}}$. The optimal results are indicated in \textbf{bold}.}
\label{result_ablation}
\centering
\resizebox{1\textwidth}{!}{
\begin{tabular}{l|lllll|lllll|lllll}
\hline
\multirow{2}{*}{Method}&\multicolumn{5}{|c|}{Upernavik}&\multicolumn{5}{|c|}{Hayes}&\multicolumn{5}{|c}{Kangerlussuaq}\\
\cline{2-16}
&MAE$\downarrow$&RMSE$\downarrow$&R$^{2}\uparrow$&SSIM$\uparrow$&PSNR$\uparrow$&MAE$\downarrow$&RMSE$\downarrow$&R$^{2}\uparrow$&SSIM$\uparrow$&PSNR$\uparrow$&MAE$\downarrow$&RMSE$\downarrow$&R$^{2}\uparrow$&SSIM$\uparrow$&PSNR$\uparrow$\\
\hline\hline
w/o $\nabla$&8.576&12.643&0.995&0.975&39.954&20.831&32.366&0.990&0.934&37.395&12.572&19.305&0.998&0.992&45.644\\
w/o $\mathcal{T}$&11.286&16.549&0.991&0.952&37.616&55.710&110.856&0.879&0.826&26.702&17.468&27.465&0.997&0.983&42.581\\
w/o $\nabla$ \& $\mathcal{T}$&18.039&28.776&0.972&0.921&32.811&73.440&128.588&0.837&0.746&25.413&27.142&43.694&0.992&0.970&38.549\\
w/o $\mathcal{L}_{\text{ref}}$&39.141&105.512&0.621&0.829&21.526&171.943&316.607&0.014&0.524&17.587&209.995&2158.767&-18.778&0.896&4.673\\
\textbf{GraphTopoNet}&\textbf{7.328}&\textbf{10.392}&\textbf{0.996}&\textbf{0.979}&\textbf{41.658}&\textbf{19.740}&\textbf{31.530}&\textbf{0.990}&\textbf{0.938}&\textbf{37.623}&\textbf{11.523}&\textbf{18.549}&\textbf{0.999}&\textbf{0.992}&\textbf{45.991}\\
\hline
\end{tabular}
}
\end{table*}

\begin{table*}[ht!]
\caption{Qualitative comparison of full-grid predictions in the Upernavik region. Each row shows the predicted bed topography, the BedMachine reference, and their difference map.}
    \label{qualitative_idw}
    \centering
    \begin{tabular}{cc}\hline
    Prediction $\mid$ BedMachine (references) $\mid$ Difference&Prediction $\mid$ BedMachine (reference) $\mid$ Difference\\
    \hline\hline
    IDW~\cite{johnston2001using,shepard1968two}&U-Net++~\cite{zhou2018unet++}\\
    \begin{minipage}[c]{0.48\textwidth}\includegraphics[width=1\textwidth]{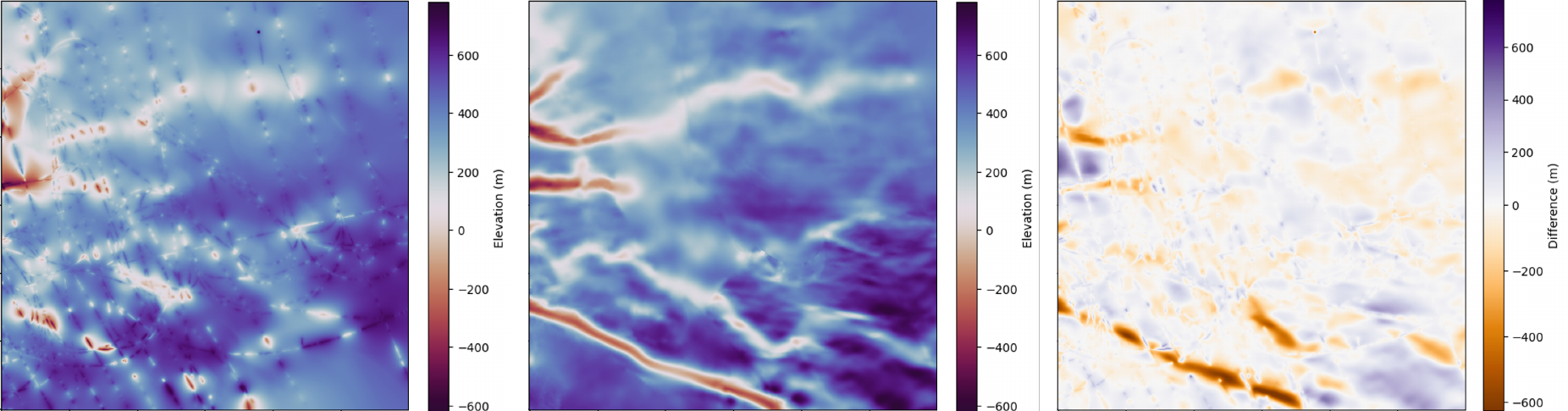}\end{minipage}&
    \begin{minipage}[c]{0.48\textwidth}\includegraphics[width=1\textwidth]{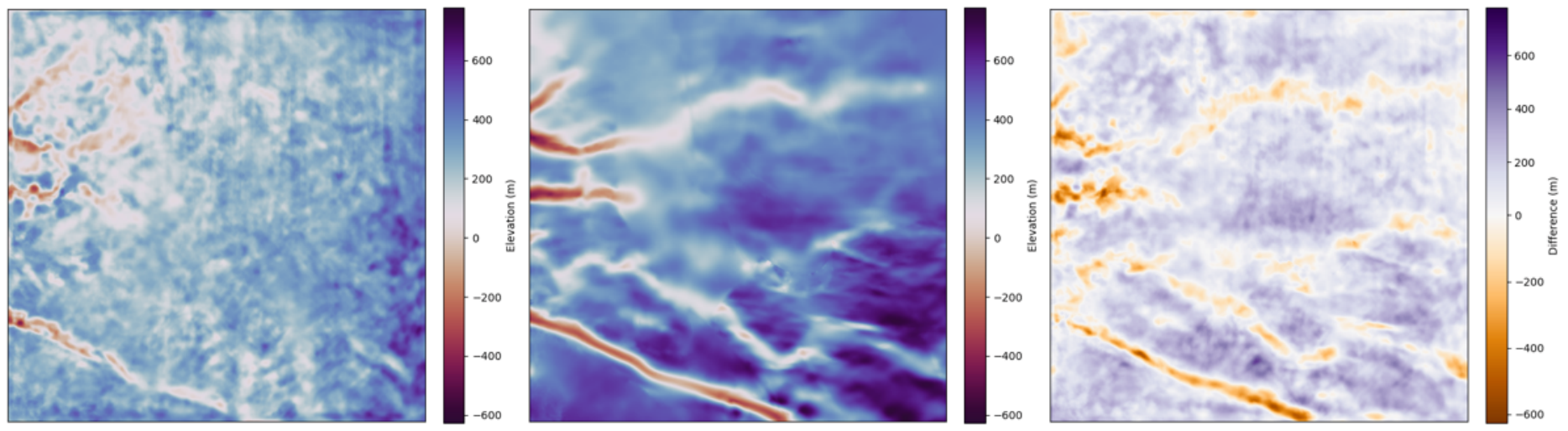}\end{minipage}\\
    Nearest Neighbor~\cite{hart1968condensed}&U-Net3+~\cite{huang2020unet}\\
    \begin{minipage}[c]{0.48\textwidth}\includegraphics[width=1\textwidth]{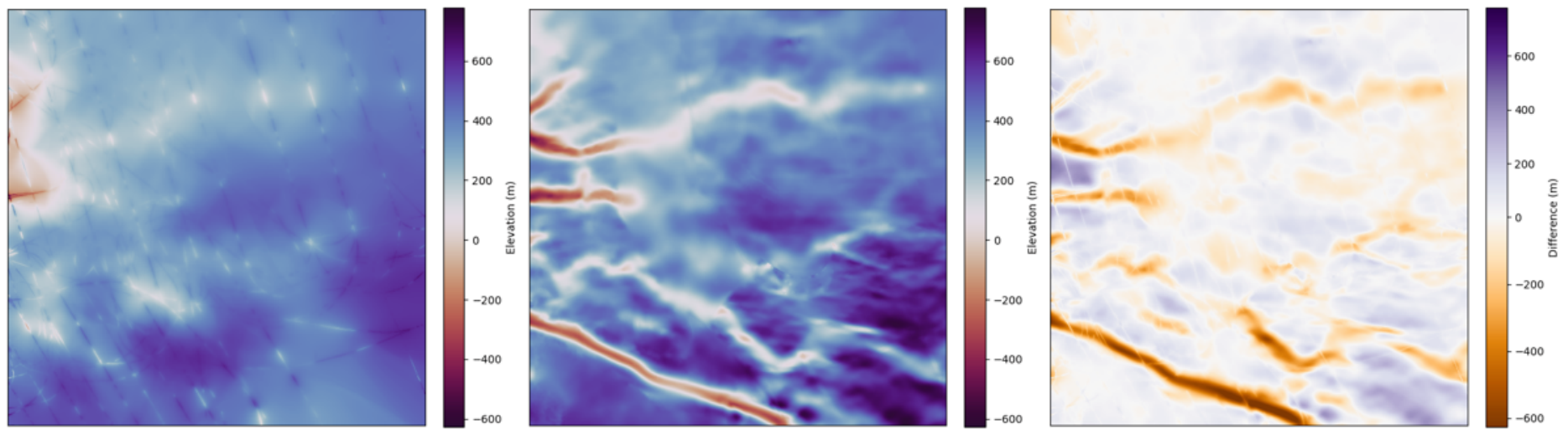}\end{minipage}&
    \begin{minipage}[c]{0.48\textwidth}\includegraphics[width=1\textwidth]{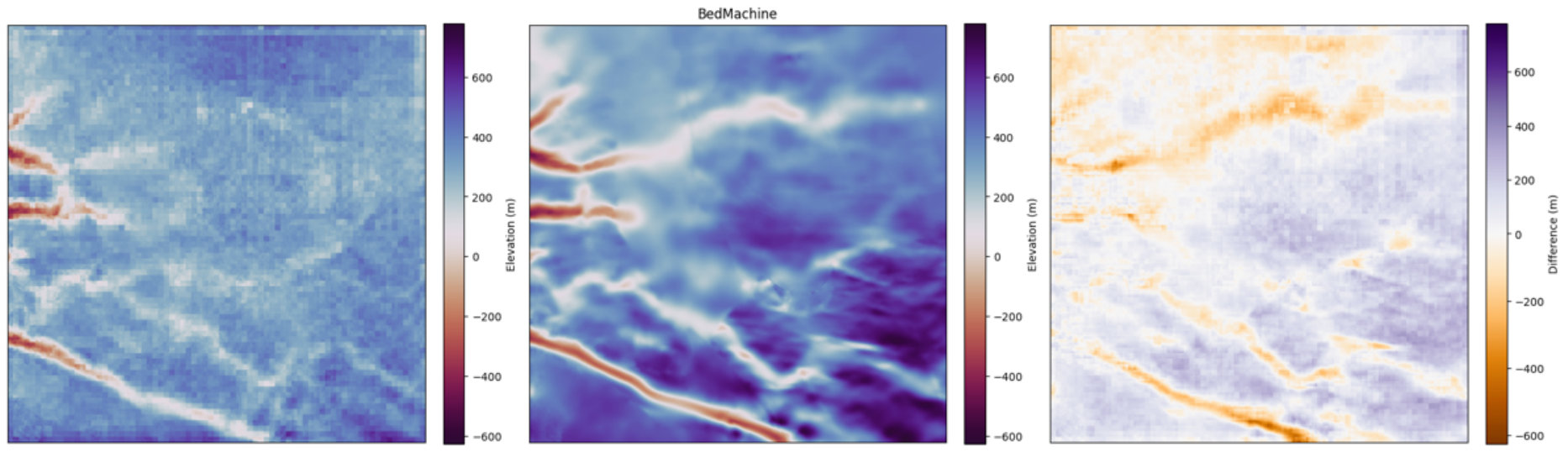}\end{minipage}\\
    GSGI~\cite{mitas1999spatial}&Att. U-Net~\cite{oktay2018attention}\\
    \begin{minipage}[c]{0.48\textwidth}\includegraphics[width=1\textwidth]{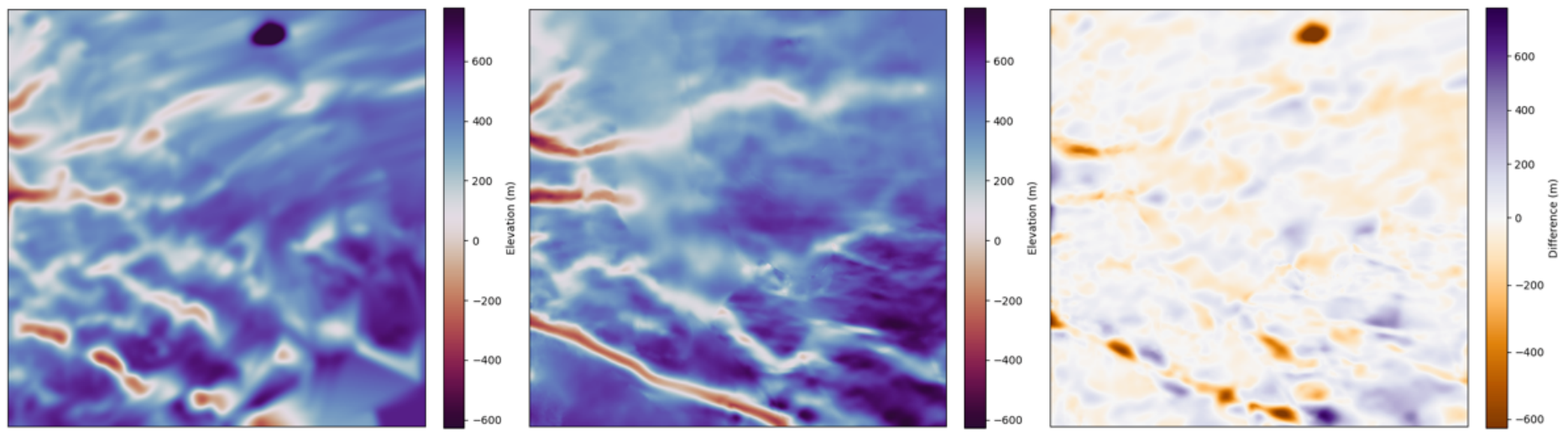}\end{minipage}&
    \begin{minipage}[c]{0.48\textwidth}\includegraphics[width=1\textwidth]{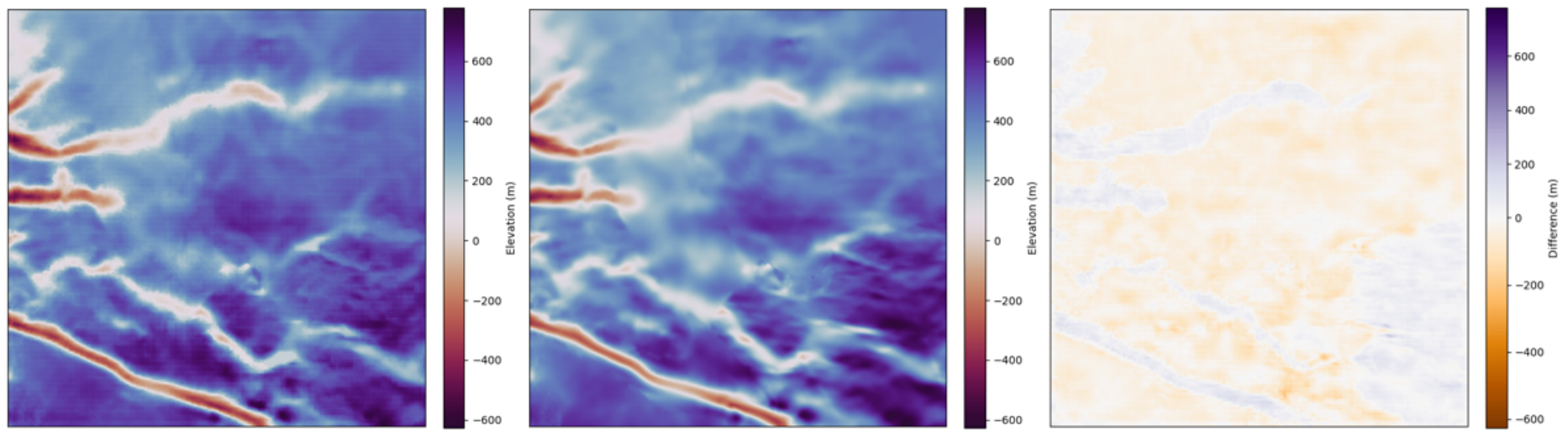}\end{minipage}\\
    MLP~\cite{schmidhuber2015deep}&ChebNet~\cite{defferrard2016convolutional}\\
    \begin{minipage}[c]{0.48\textwidth}\includegraphics[width=1\textwidth]{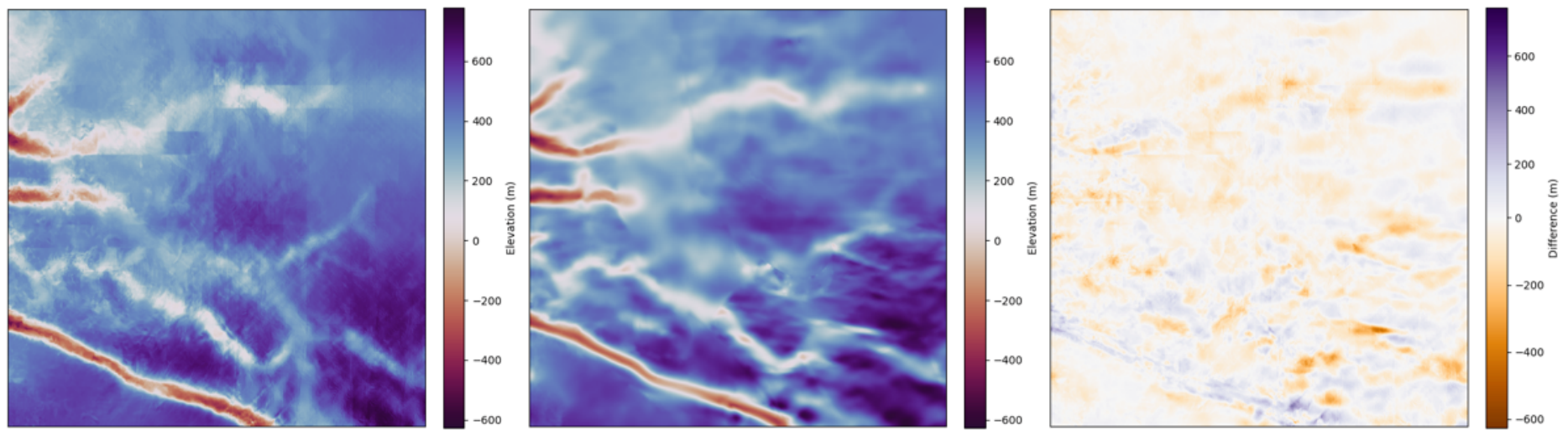}\end{minipage}&
    \begin{minipage}[c]{0.48\textwidth}\includegraphics[width=1\textwidth]{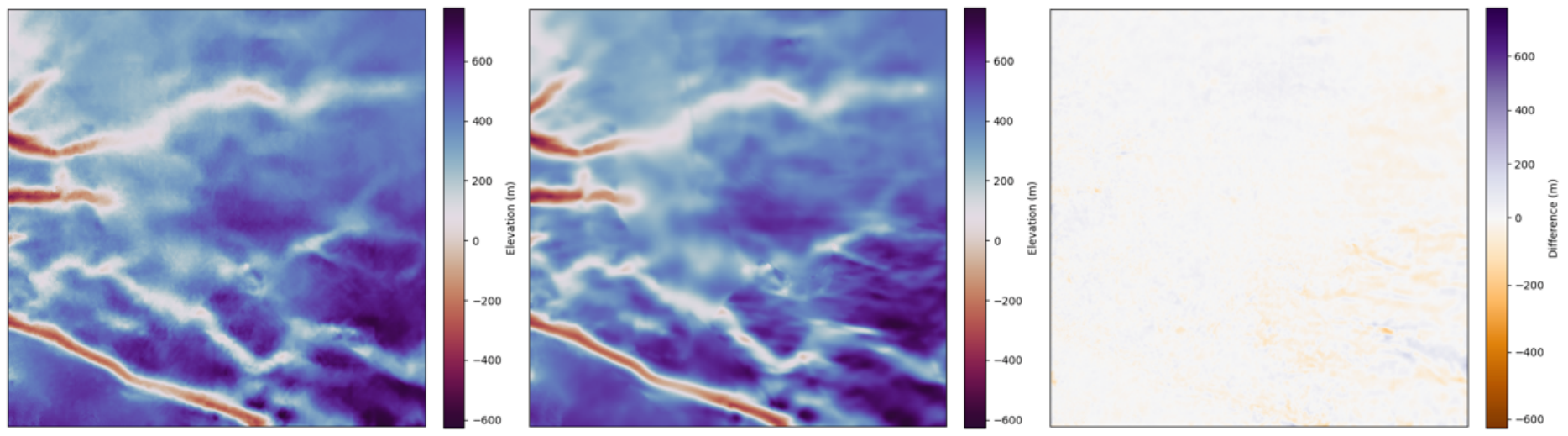}\end{minipage}\\
    MLP-PE~\cite{li2021learnable}&GraphSAGE~\cite{hamilton2017inductive}\\
    \begin{minipage}[c]{0.48\textwidth}\includegraphics[width=1\textwidth]{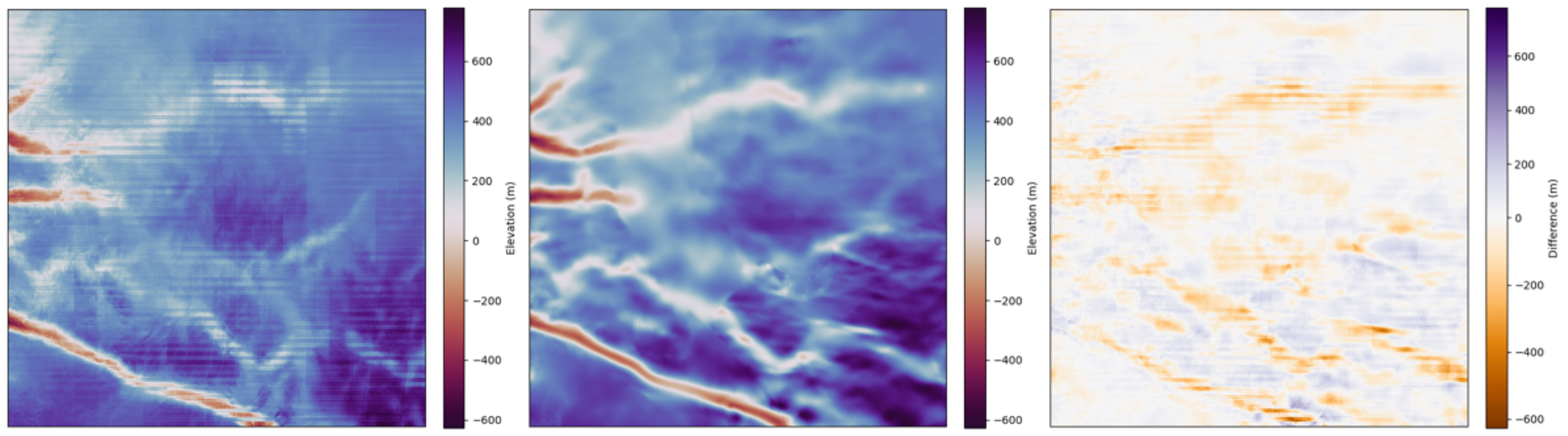}\end{minipage}&
    \begin{minipage}[c]{0.48\textwidth}\includegraphics[width=1\textwidth]{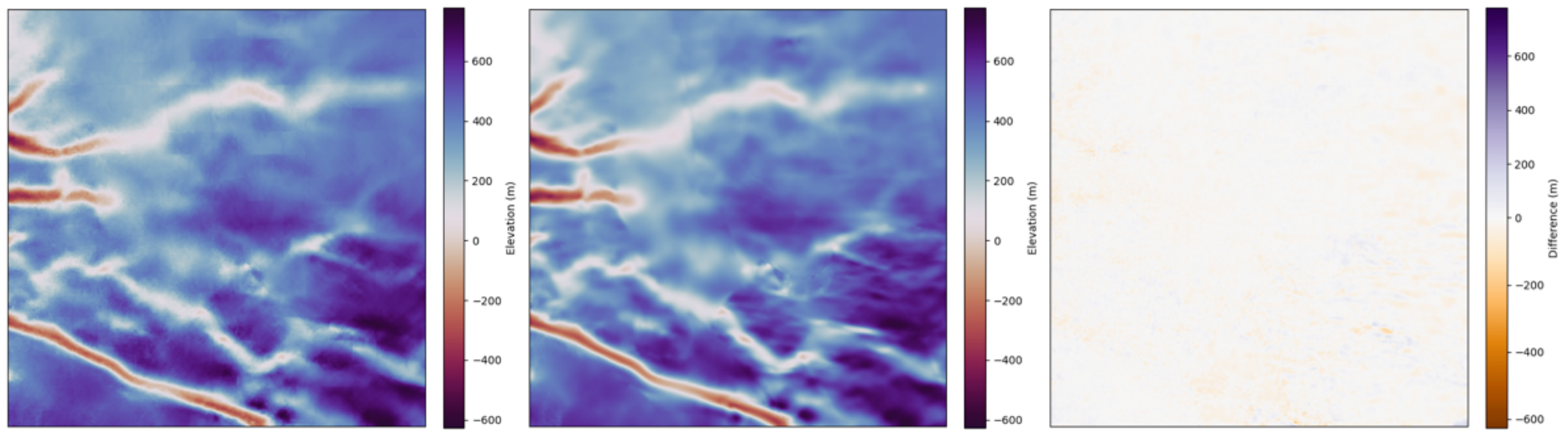}\end{minipage}\\
    MLP-FE~\cite{tancik2020fourier}&GAT~\cite{veličković2018graph}\\
    \begin{minipage}[c]{0.48\textwidth}\includegraphics[width=1\textwidth]{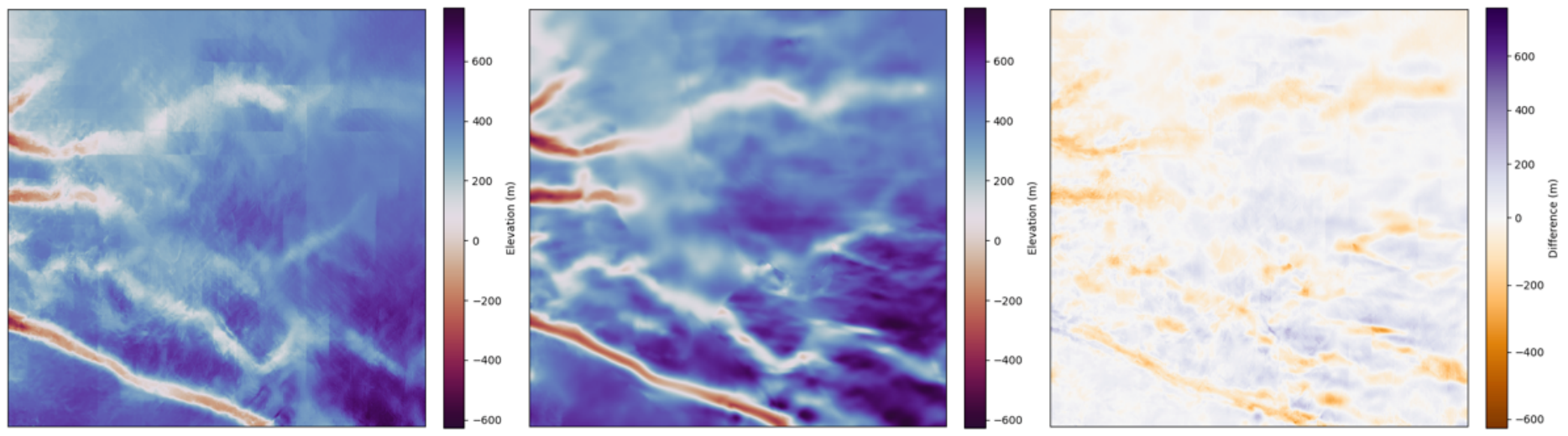}\end{minipage}&
    \begin{minipage}[c]{0.48\textwidth}\includegraphics[width=1\textwidth]{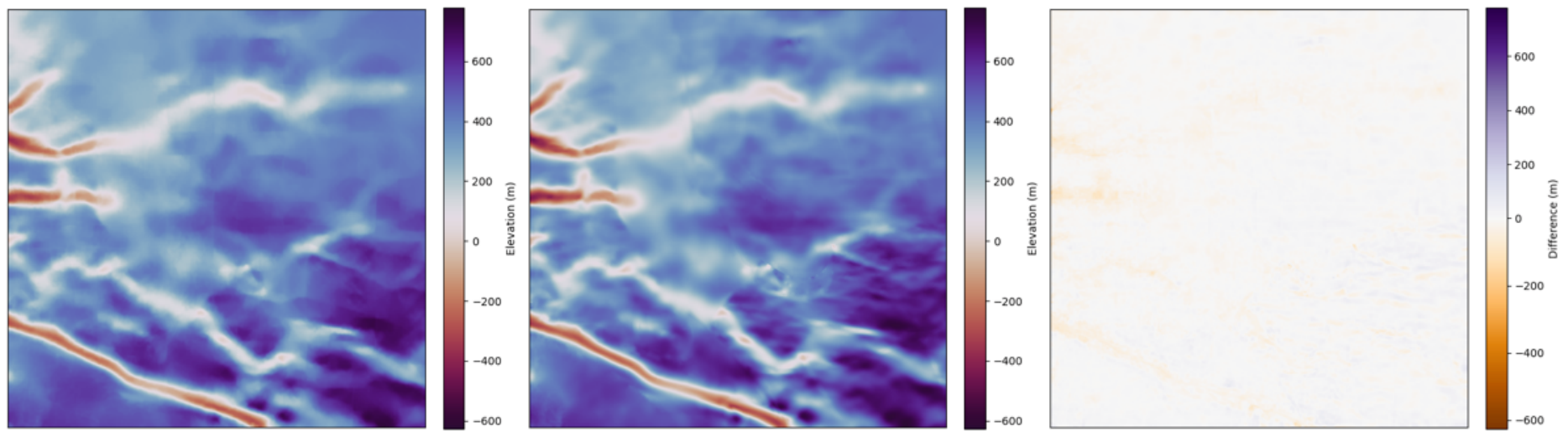}\end{minipage}\\
    U-Net~\cite{ronneberger2015u}&\textbf{GraphTopoNet}\\
    \begin{minipage}[c]{0.48\textwidth}\includegraphics[width=1\textwidth]{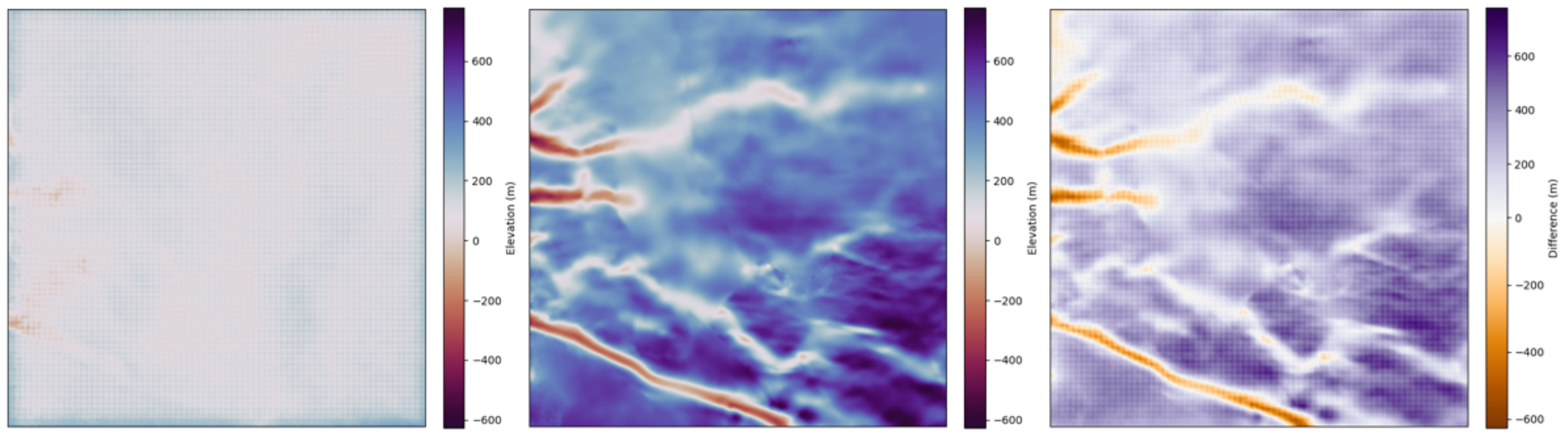}\end{minipage}&
    \begin{minipage}[c]{0.48\textwidth}\includegraphics[width=1\textwidth]{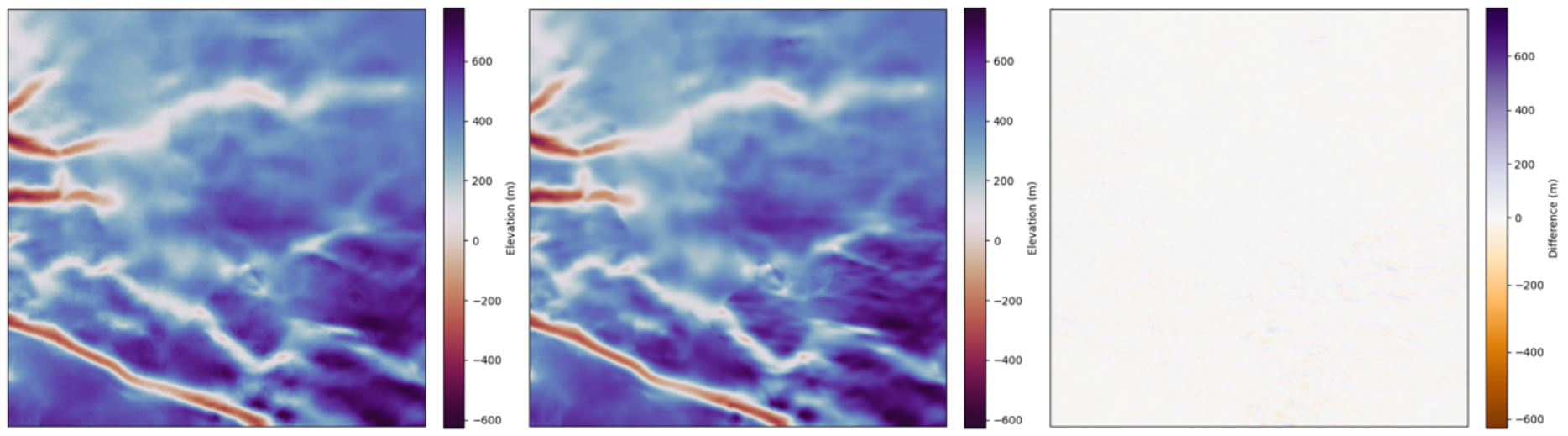}\end{minipage}\\
    &\\
    \hline
    \end{tabular} 
\end{table*}

\section{Results}
In this section, we evaluate the performance of our proposed method, \texttt{GraphTopoNet}, against a variety of baseline approaches for bed topography prediction. We used the experiment environment as follow: PC with Linux OS, Intel Xeon 2.4GHz processor, 256GB RAM, and GPU RTX A4500.

\subsection{Quantitative Evaluation}
The quantitative results in Table~\ref{result_quantitative} show distinct performance trends across method categories. \texttt{Interpolation-based methods} such as IDW, Nearest Neighbor, and GSGI demonstrate relatively poor performance, especially in regions with sparse radar coverage (e.g., Hayes), due to their reliance on local spatial proximity without learning underlying physical or spatial patterns. While GSGI performs slightly better among the three—especially in SSIM and R$^2$—it still lags significantly behind learning-based approaches.

\texttt{Machine learning models} (MLP, MLP-PE, MLP-FE) show moderate improvements over interpolation baselines. These models benefit from data-driven learning but remain limited in capturing spatial context due to their fully connected architecture. The use of positional encoding and feature augmentation (PE, FE) further boosts performance but with diminishing returns in complex terrains like Kangerlussuaq. \texttt{Deep learning models} such as U-Net variants (U-Net, U-Net++, U-Net3+, Att. U-Net) produce mixed results. The basic U-Net and its variants (U-Net++ and U-Net3+) fail to generalize well in low-data regions, with large errors and even negative R$^2$ values in some cases. In contrast, the Attention U-Net shows strong improvements across all metrics, indicating that spatial attention mechanisms help refine predictions by focusing on relevant features.

\texttt{Graph-based models}—ChebNet, GraphSAGE, and GAT—outperform all previous categories by a wide margin, showcasing their ability to leverage non-Euclidean spatial structures and incorporate connectivity-aware learning. Notably, \texttt{GraphTopoNet}, our proposed model, achieves the best performance across all sub-regions and evaluation metrics. It consistently produces lower errors (MAE and RMSE), higher fidelity (SSIM and PSNR), and excellent predictive fit (R$^2$ nearing 1), validating the effectiveness of graph learning combined with uncertainty and confidence-aware modeling.

\subsection{Qualitative Analysis}
To complement the quantitative evaluation, Figure~\ref{qualitative_idw} presents a qualitative comparison of full-grid bed topography predictions across methods for the Upernavik region. Each subfigure displays the predicted topography, the corresponding BedMachine reference, and the residual difference. Interpolation methods such as IDW and Nearest Neighbor exhibit pronounced artifacts and discontinuities, particularly near unobserved regions, reflecting their limited generalization capacity. Classical MLP-based methods demonstrate smoother reconstructions but still fail to capture finer structural variations compared to deep learning counterparts. Among deep architectures, U-Net variants struggle with over-smoothing and boundary distortion, while the Attention U-Net shows improved delineation of glacial channels. Graph-based models, particularly GAT and GraphSAGE, achieve visually consistent and geophysically plausible reconstructions with low residuals. Our proposed GraphTopoNet clearly outperforms all baselines, producing the most faithful bed topography with minimal deviation from the reference, especially in poorly observed areas—highlighting its robustness and effectiveness in uncertainty-aware spatial learning.

\subsection{Ablation Study}
To assess the contribution of key model components and input features, we conduct an ablation study by systematically disabling specific elements in the GraphTopoNet pipeline. Table~\ref{result_ablation} summarizes the performance degradation observed when the following components are excluded: 
\begin{itemize}
\item \textbf{w/o $\nabla$}: Excludes gradient-based covariates derived from the surface input fields. Gradients capture local spatial variations, and their removal leads to a noticeable drop in accuracy across all metrics, particularly in Upernavik and Kangerlussuaq.
\item \textbf{w/o $\mathcal{T}$}: Removes the polynomial trend surface features, which model coarse-scale spatial variation. This exclusion significantly degrades performance—especially in Hayes—suggesting that trend information is critical in regions with more complex bed geometry.
\item \textbf{w/o $\nabla$ \& $\mathcal{T}$}: Excludes both gradient and trend surface features. This combined removal results in the most severe performance degradation among feature ablations, highlighting the complementary role of fine-scale (gradients) and large-scale (trends) surface representations. 
\item \textbf{w/o $\mathcal{L}_{\text{ref}}$}: Disables the auxiliary loss term supervised by the full-grid BedMachine reference. This variant exhibits a drastic drop in performance across all metrics and regions, particularly in unobserved areas, confirming that $\mathcal{L}_{\text{ref}}$ provides critical regularization to guide the model beyond radar-labeled points.
\end{itemize}

The full model, \texttt{GraphTopoNet}, achieves the best overall performance, demonstrating the importance of integrating both gradient and trend features alongside dual-source supervision. The results highlight how each architectural and supervision component contributes to spatial accuracy, structural consistency (SSIM), and overall topographic fidelity (PSNR).

\subsection{Evaluating Spatial Extrapolation via Grid-Based Partitioning}
To evaluate the model's ability to generalize to unseen spatial regions, we adopt a grid-based slicing strategy inspired by prior work on spatial transfer learning~\cite{beery2018recognition,meyer2019importance}. Specifically, the spatial domain is divided into 30 horizontal bands of equal width, with alternating slices assigned to training (odd-numbered: 1, 3, 5, ...) and testing (even-numbered: 2, 4, 6, ...). This approach simulates real-world deployment scenarios where radar observations are sparsely distributed, requiring the model to extrapolate across gaps in spatial coverage and predict bed topography in unobserved transects.

\begin{table}[ht!]
\caption{Performance comparison of GraphTopoNet and baseline GNN models under grid-based spatial partitioning in the Upernavik sub-region. Models are evaluated on alternating horizontal slices held out from training to assess spatial extrapolation.}
    \label{grid_based_partition}
    \centering
    \resizebox{0.5\textwidth}{!}{
    \begin{tabular}{l|lllll}
    \hline 
    \multirow{2}{*}{Method}&\multicolumn{5}{c}{Upernavik}\\
    \cline{2-6}\\    
    &MAE$\downarrow$&RMSE$\downarrow$&R$^{2}\uparrow$&SSIM$\uparrow$&PSNR$\uparrow$\\
    \hline\hline
    GAT~\cite{veličković2018graph}&56.640&79.220&0.787&0.796&24.015\\    GraphSAGE~\cite{hamilton2017inductive}&13.504&18.834&0.988&0.942&36.493\\    \textbf{GraphTopoNet}&9.587&15.084&0.992&0.965&38.421\\
    \hline
    \end{tabular}   
    }
\end{table}

Table~\ref{grid_based_partition} reports the results for the Upernavik sub-region under this grid-based evaluation. As expected, all models exhibit performance degradation compared to the random-split results shown in Table~\ref{result_quantitative}, confirming the increased difficulty of spatial extrapolation. Notably, \texttt{GraphTopoNet} retains strong predictive performance despite the reduced spatial overlap between training and test regions, outperforming GAT and GraphSAGE across all metrics. In particular, it achieves an MAE of 9.59 m and an R$^2$ of 0.992—substantially higher than the GAT baseline, which suffers from a sharp drop in predictive accuracy (MAE = 56.64 m, R$^2$ = 0.787).

These results demonstrate that while general-purpose graph models like GAT and GraphSAGE are effective under i.i.d. training conditions, their extrapolation capacity is limited when applied to spatially disjoint regions. In contrast, GraphTopoNet’s integration of spatial priors (gradients, trend surfaces) and dual-source supervision (radar + BedMachine) enhances its ability to make structurally coherent and topographically plausible predictions in unseen areas.

\section{Discussion and Limitations}
While GraphTopoNet demonstrates strong performance across three Greenland sub-regions, several limitations remain. First, the framework currently depends on BedMachine as a fallback supervision source. Although widely used in glaciology, this dataset itself contains biases in slow-flowing regions, which may propagate into predictions. Second, our evaluation is limited to three sub-regions of Greenland; a natural next step is to extend the framework to the entire Greenland Ice Sheet before exploring transferability to Antarctica or other sparse geophysical domains. Third, while patch-based training ensures scalability to 600$\times$600 grids on modern GPUs, applying the framework at continental scale requires distributed training strategies. 

Despite these limitations, the work highlights a key lesson for applied spatial data mining: integrating heterogeneous supervision sources with uncertainty-awareness enables models to remain useful even under imperfect data availability. For government and scientific stakeholders, this provides a practical path toward improving operational products (e.g., subglacial maps) that inform global climate projections and policy. Future work will incorporate physics-informed constraints and temporal surface features to further enhance robustness and applicability.

\section{Conclusion}
In this work, we introduced \textbf{GraphTopoNet}, a graph-based deep learning framework for high-resolution spatial prediction under sparse and uncertain supervision. By constructing spatial graphs from surface-derived covariates and incorporating both gradient and polynomial trend surface augmentations, GraphTopoNet effectively captures local and global spatial dependencies. We proposed a hybrid loss function that integrates radar confidence-weighted supervision, full-coverage reference guidance, and epistemic uncertainty regularization via Monte Carlo Dropout. Extensive experiments across three Greenland sub-regions demonstrate that GraphTopoNet consistently outperforms interpolation, convolutional, and graph-based baselines in both random and spatially disjoint evaluation settings. The model generates accurate, structurally coherent, and uncertainty-aware predictions, particularly in regions with limited observational coverage. Future work includes extending the framework with additional physical constraints (e.g., mass conservation or ice dynamics) to enhance physical realism. Incorporating temporal surface features and applying the model in a spatiotemporal context may also support time-resolved prediction of bed evolution. Finally, evaluating GraphTopoNet on other sparse spatial domains (e.g., hydrology or air quality) could demonstrate its broader utility in geospatial data mining. 

\section{Acknowledgement}
This research has been funded by the NSF HDR Institute for Harnessing Data and Model Revolution in the Polar Regions (iHARP), Award \#2118285.

\bibliographystyle{IEEEtran}
\bibliography{mybib}

\end{document}